%% file: main.tex
\documentclass[10pt,twocolumn,letterpaper]{article}

\input{preamble}

\author{
Xiang Zhang$^{*,1,2,\dagger}$
\quad Sohyun Yoo$^{*,1}$
\quad Hongrui Wu$^{*,1,\ddagger}$
\quad Chuan Li$^2$
\quad Jianwen Xie$^2$
\quad Zhuowen Tu$^1$ \\
$^1$UC San Diego \quad $^2$Lambda, Inc. \\
{\small $^*$equal contribution}
}

\begin{document}
\maketitle
\input{sec/0_abstract}
\customfootnotetext{$\dagger$}{Work partially done during internship at Lambda.}
\customfootnotetext{$\ddagger$}{H. Wu contributed to the work during internship at UC San Diego.}
\customfootnotetext{}{Project page: \url{https://mlpc-ucsd.github.io/PixARMesh}}
\input{sec/1_intro}
\input{sec/2_related_work}
\input{sec/3_method}
\input{sec/4_experiments}
\input{sec/5_conclusion}
\input{sec/acknowledgment}

{
    \small
    \bibliographystyle{ieeenat_fullname}
    \bibliography{main}
}

\iftoggle{arxiv}{
    \clearpage
    \appendix
    
    \counterwithin{figure}{section}
    \counterwithin{table}{section}
    \counterwithin{equation}{section}
    \setcounter{table}{0}
    \setcounter{figure}{0}
    \setcounter{equation}{0}
    \section{Appendix}
    
    \let\origsection\section
    \let\origsubsection\subsection
    \let\origsubsubsection\subsubsection
    \let\section\subsection
    \let\subsection\subsubsection
    
    \input{sec/supplementary}

}{}
\end{document}

%% file: preamble.tex
\usepackage{etoolbox}
\newtoggle{arxiv}
\toggletrue{arxiv}
\iftoggle{arxiv}{
\usepackage[pagenumbers]{cvpr}
}{
\usepackage{cvpr}
}

\usepackage{float}
\usepackage{booktabs}
\usepackage{comment}
\usepackage{multirow}
\usepackage{lipsum}
\usepackage{xspace}
\usepackage{siunitx}
\usepackage{pifont}
\definecolor{cvprblue}{rgb}{0.21,0.49,0.74}
\usepackage[pagebackref,breaklinks,colorlinks,allcolors=cvprblue]{hyperref}
\usepackage[table]{xcolor}
\usepackage{colortbl}
\usepackage[accsupp]{axessibility}

\newcommand{\cmark}{\ding{51}}%

\newcommand{\OURS}{PixARMesh\xspace}

\title{\OURS: Autoregressive Mesh-Native Single-View Scene Reconstruction}

\newcommand{\customfootnotetext}[2]{%
  \begingroup
  \renewcommand{\thefootnote}{#1}%
  \footnotetext{#2}%
  \endgroup
}

\makeatletter
\renewcommand{\paragraph}{%
  \@startsection{paragraph}{4}{\z@}%
  {0.3ex plus 0.1ex minus 0.1ex}
  {-1em}
  {\normalfont\normalsize\bfseries}%
}
\makeatother

%% file: sec/0_abstract.tex
\begin{abstract}

We introduce \OURS, a method to autoregressively reconstruct complete 3D indoor scene meshes directly from a single RGB image. Unlike prior methods that rely on implicit signed distance fields and post-hoc layout optimization, \OURS jointly predicts object layout and geometry within a unified model, producing coherent and artist-ready meshes in a single forward pass. Building on recent advances in mesh generative models, we augment a point-cloud encoder with pixel-aligned image features and global scene context via cross-attention, enabling accurate spatial reasoning from a single image. Scenes are generated autoregressively from a unified token stream containing context, pose, and mesh, yielding compact meshes with high-fidelity geometry. Experiments on synthetic and real-world datasets show that \OURS achieves state-of-the-art reconstruction quality while producing lightweight, high-quality meshes ready for downstream applications.

\end{abstract}

%% file: sec/1_intro.tex
\section{Introduction}
\label{sec:intro}

Reconstructing a complete 3D scene from a single RGB image is a long-standing and fundamentally ill-posed problem. A single viewpoint provides only partial, depth-ambiguous observations of objects, while large portions of the scene remain occluded or unobserved. Recovering accurate object shapes and coherent spatial layouts therefore requires strong priors about indoor scenes and plausible object structures.

Earlier methods~\cite{dahnert2021panoptic,chu2023buol,zhang2023uni} reconstruct the entire scene holistically by back-projecting image features into 3D volumes and predicting a scene-level signed distance field (SDF) using an encoder-decoder architecture. While these approaches bypass explicit layout estimation, they remain fundamentally constrained by the spatial resolution of volumetric grids and the limited expressiveness of feed-forward decoders. As a result, they struggle to produce high-quality geometry and lack the generative flexibility and generalization capability needed for complex real-world scenes.

\begin{figure}[!t]
    \centering
    \includegraphics[width=\linewidth,trim=0 10pt 0 0,clip]{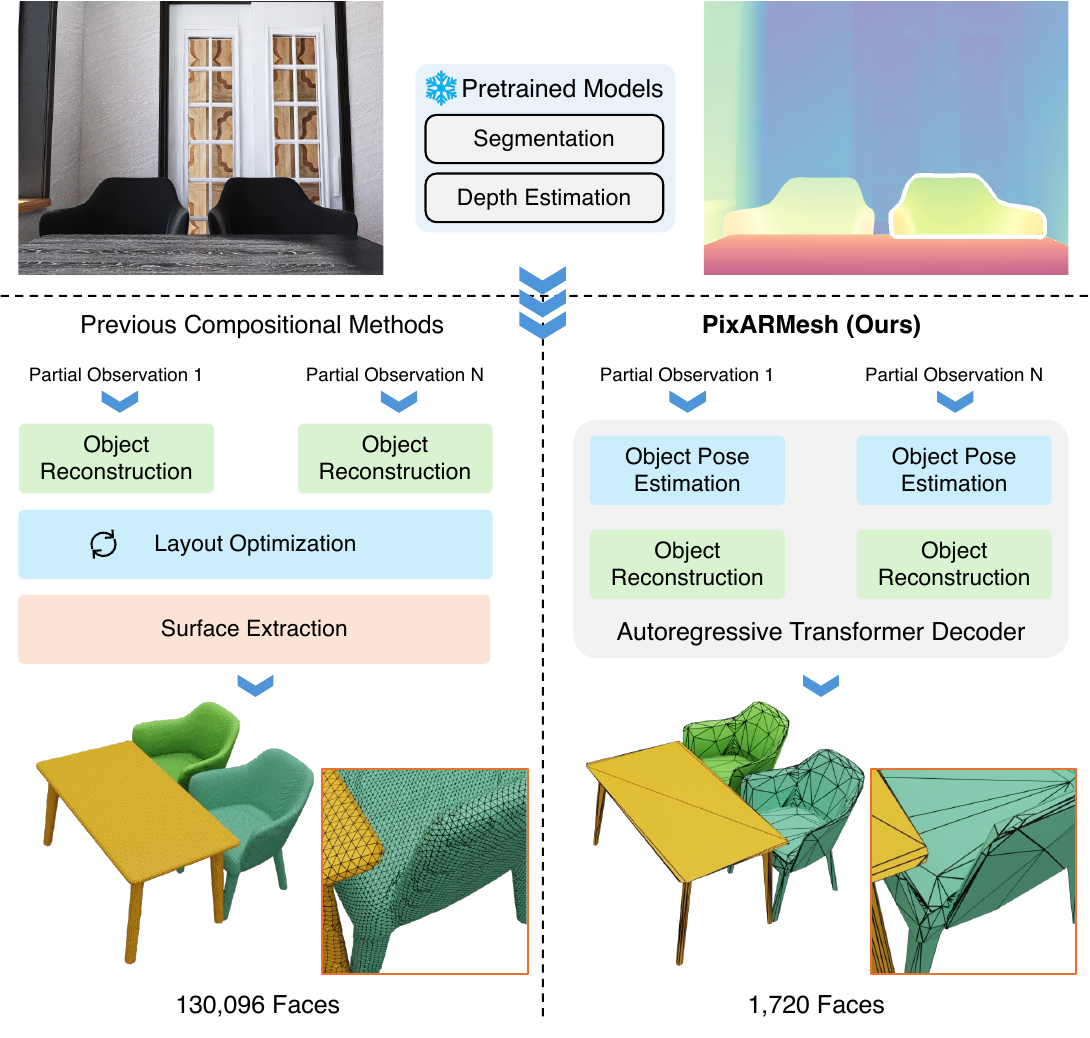}
    \caption{Comparison of \OURS with recent compositional scene reconstruction methods. \OURS predicts object poses and reconstructs native meshes in a single autoregressive decoding process, without relying on SDF-based surface extraction or layout optimization, producing compact and 
    artist-ready mesh outputs.}
    \label{fig:teaser}
    \vspace{-3pt}
\end{figure}

Recently, the compositional generation paradigm has gained significant attention, driven by advances in large-scale object-level reconstruction models~\cite{hong2023lrm,jun2023shap,liu2023zero,liu2023one,liu2024one,zhang2024clay}. Since these models are typically pre-trained on clean, unoccluded object images, existing pipelines~\cite{zhou2024zero,dogaru2024generalizable} require an inpainting or amodal completion stage to recover occluded regions before passing object crops to the reconstruction network. To assemble the reconstructed instances into a coherent scene, they further rely on optimization-based layout estimation, often formulated as point-cloud matching, which can be prone to local minima. Recent works such as DepR~\cite{Zhao_2025_ICCV_DepR} mitigate the need for inpainting by conditioning generation directly on partial observations, while MIDI~\cite{huang2025midi} eliminates layout optimization by predicting each instance directly in normalized scene coordinates. Although these methods generally achieve higher reconstruction fidelity, their dependence on SDF-based representations introduces additional complexity in surface extraction and often yields overly smooth, high-face-count meshes that deviate from artist-ready geometry.

Meanwhile, there is steady progress in object-level mesh generative models~\cite{siddiqui2024meshgpt,chen2024meshxl,chen2024meshanything,chen2025meshanything,tang2024edgerunner,weng2025scaling,zhang2025vertexregen,lei2025armesh}, where artist-like mesh sequences are directly predicted by an autoregressive Transformer decoder, eliminating the need for iso-surface extraction. However, despite these advances, autoregressive mesh generators remain limited to object-level outputs, and no existing scene reconstruction pipeline leverages their native, artist-ready mesh representations. This gap motivates the integration of strong partial observations with mesh-level generative priors for scene-level reconstruction.

To bridge this gap, we propose \OURS, a framework built on top of pre-trained object-level autoregressive mesh generative models such as EdgeRunner~\cite{tang2024edgerunner} and BPT~\cite{weng2025scaling}, introducing a new paradigm for single-view scene reconstruction using native, artist-ready mesh representations. This architectural shift allows us to replace complex optimization loops with a unified, feed-forward autoregressive sequence (see \cref{fig:teaser}), ensuring coherent scene composition and high-fidelity geometry. To leverage the limited geometric cues available in depth-back-projected point clouds, we fuse pixel-aligned image features into the point-cloud encoder, injecting appearance cues on top of partial geometry. To further enhance scene-level understanding, we incorporate cross-attention between each object’s point-cloud features and a global scene point cloud, enabling context-aware reconstruction under heavy occlusion. Finally, we utilize the coordinate vocabulary of existing mesh generative models to tokenize scene composition, allowing \OURS to jointly predict object poses and meshes within a single feed-forward autoregressive sequence. We validate \OURS on synthetic 3D-FRONT~\cite{3DFRONT} and real-world images, demonstrating that it produces high-quality, artist-ready meshes with coherent layouts and strong reconstruction performance.

Our main contributions are summarized as follows:
\begin{itemize}
    \item We present the first framework that performs single-view scene reconstruction \textit{directly}, \textit{autoregressively} in mesh space, avoiding SDF-based decoding and surface extraction while producing high-quality, artist-ready outputs.
    \item We repurpose recent object-level mesh generative models by incorporating \textit{pixel-aligned image features} and \textit{global scene context} into the point-cloud encoder, enabling context-aware pose and geometry generation from a single image.
    \item We jointly predict object poses and meshes in a single feed-forward autoregressive manner, achieving coherent scene composition without post-hoc layout optimization. Extensive experiments demonstrate \OURS achieves state-of-the-art reconstruction performance.
\end{itemize}

%% file: sec/2_related_work.tex
\section{Related Work}

\paragraph{3D Scene Reconstruction from a Single Image} Single-view reconstruction is inherently ill-posed due to scale ambiguity, occlusion, and incomplete geometric cues, often requiring depth or shape priors from large-scale pre-trained models. Early holistic approaches such as Panoptic3D~\cite{liu2021towards}, PanoRe~\cite{dahnert2021panoptic}, Uni-3D~\cite{zhang2023uni}, and BUOL~\cite{chu2023buol} reconstruct an entire scene using feed-forward encoder-decoder architectures applied to back-projected feature volumes. While these methods do not require explicit layout estimation, they are constrained by limited spatial resolution and exhibit poor generalization and generative capability.

Recent research has shifted toward compositional generation frameworks that decompose a scene into individual instances and reconstruct them before composing the final scene. Some approaches~\cite{kuo2020mask2cad,huang2025litereality} rely on shape retrieval and procedural assembly, while others benefit from advances in object-level generative models. For example, Gen3DSR~\cite{dogaru2024generalizable} and DeepPriorAssembly~\cite{zhou2024zero} perform image inpainting to complete occluded regions before feeding the recovered object crops into pre-trained object reconstruction models~\cite{jun2023shap,liu2023zero,zhang2024clay,xu2024bayesian}. DepR~\cite{Zhao_2025_ICCV_DepR} instead generates shapes conditioned on partial image observations using a depth-guided diffusion model. These methods rely on post-hoc, optimization-based layout estimation to compose reconstructed instances back into a scene, which can be susceptible to local minima and spatial misalignment. MIDI~\cite{huang2025midi} alleviates this limitation by generating all instances within a normalized scene space, thereby avoiding explicit pose estimation. Despite these advances, most existing approaches operate on signed distance fields (SDFs) and require iso-surface extraction via marching cubes~\cite{lorensen1998marching}, often producing densely tessellated and overly smooth meshes that hinder geometry-based applications such as editing. Our work addresses these limitations by unifying object layout prediction and mesh-native reconstruction into a single autoregressive sequence.

\begin{figure*}[!ht]
    \centering
    \includegraphics[width=\linewidth]{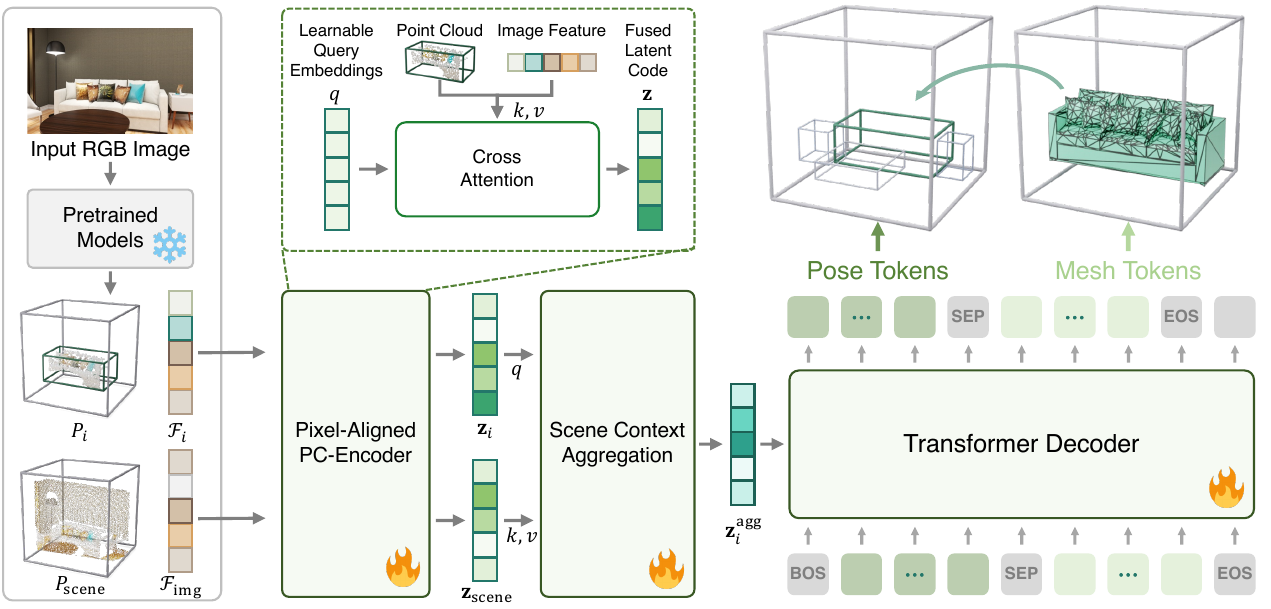}
    \caption{Pipeline overview. Given an RGB image, we use pretrained models to extract the depth point cloud and image features for both the target object $i$ and the global scene. These local and global cues are fed into the Pixel-Aligned PC-Encoder to produce the fused latent code, which is then aggregated into a single latent vector via cross-attention. This latent vector conditions the Transformer Decoder, which predicts the object's pose followed by its mesh token sequence.}
    \label{fig:pipeline}
\end{figure*}

\paragraph{Native Mesh Generation} Generating 3D shapes directly in native, artist-like meshes has long been attractive for their compactness, editability, and compatibility with downstream graphics applications. Early methods rely on structured primitives such as surface patches~\cite{groueix2018papier}, deformable ellipsoids~\cite{wang2018pixel2mesh}, mesh graphs~\cite{dai2019scan2mesh}, and binary space partitioning~\cite{chen2020bsp}, but they typically impose strong geometric priors and offer limited topological flexibility. More recently, PolyDiff~\cite{alliegro2023polydiff} applies discrete diffusion to synthesize meshes, while PolyGen~\cite{nash2020polygen} introduces an autoregressive framework that predicts vertices and faces using two coordinated Transformer decoders.

Subsequent approaches move to a single-sequence formulation. MeshGPT~\cite{siddiqui2024meshgpt} employs a Transformer over VQ-VAE-quantized mesh tokens, and MeshAnything~\cite{chen2024meshanything} extends it with shape-conditional generation. MeshXL~\cite{chen2024meshxl} further simplifies the process by operating directly in quantized coordinate space, removing the need for a VQ-VAE but at the cost of longer token sequences. To improve scalability, recent studies propose compressive tokenization strategies that exploit face adjacency~\cite{chen2025meshanything,lionar2025treemeshgpt,tang2024edgerunner,weng2025scaling}. Meshtron~\cite{hao2024meshtron} follows MeshXL tokenization but introduces an Hourglass Transformer~\cite{nawrot2022hierarchical} to internally compress long sequences.

Others explore complementary directions for improving mesh generation quality and controllability. DeepMesh~\cite{zhao2025deepmesh} and Mesh-RFT~\cite{liu2025mesh} incorporate reinforcement learning to align mesh generation with aesthetic or human preferences. PivotMesh~\cite{weng2025pivotmesh} generates pivot vertices as coarse structural guidance for subsequent mesh generation, while VertexRegen~\cite{zhang2025vertexregen} and ARMesh~\cite{lei2025armesh} advance the coarse-to-fine generation paradigm by progressively increasing geometric detail. Building on mesh generative models with strong compression and scalability, such as EdgeRunner~\cite{tang2024edgerunner} and BPT~\cite{weng2025scaling}, our work extends these advances to scene-level reconstruction with artist-like meshes.

%% file: sec/3_method.tex
\section{Method}

We provide an overview of our framework in \cref{fig:pipeline}, which consumes depth-derived point clouds from off-the-shelf perception models and performs autoregressive scene reconstruction. We first introduce the problem setup in \cref{sec:preliminary}, then describe how we adapt point-cloud encoders from object-level mesh generative models to operate at the scene level. Finally, we detail our tokenization scheme in \cref{sec:tokenization} and our training strategy in \cref{sec:training}.

\subsection{Preliminary}\label{sec:preliminary}

The goal of single-view scene reconstruction is to recover the 3D geometry and spatial configuration of a scene from a single RGB image. Following the compositional paradigm used in prior work such as DepR~\cite{Zhao_2025_ICCV_DepR} and DeepPriorAssembly~\cite{zhou2024zero}, we focus on reconstructing only foreground object instances (\eg furniture in indoor scenes) and disregard large planar background structures such as walls and floors.

We introduce \OURS{}, an end-to-end framework that jointly predicts the shape and scene-level pose of each object instance, producing a complete scene where all objects are represented using native, artist-ready meshes rather than implicit SDFs.

Given an input RGB image $I \in \mathbb{R}^{H \times W \times 3}$, we first extract depth $D$, instance segmentation masks $\mathcal{M} = \{M_i\}_{i=1}^N$, and image features $\mathcal{F}_\mathrm{img}$ using off-the-shelf models. The depth map is back-projected using the camera intrinsics $K$ to obtain a raw scene point cloud $P_\mathrm{scene}$. Applying the instance masks yields per-object point clouds $\mathcal{P} = \{P_i\}_{i=1}^N$ where $P_i = P_\mathrm{scene} \odot M_i$, which capture only the visible portions of each object in global camera coordinates.

Unlike previous compositional methods that reconstruct object shapes first and resolve their spatial layout afterward, we unify both tasks in a single autoregressive feed-forward architecture. For each instance $i$, the model $F_\mathrm{AR}$ predicts both its scene-level pose $T_i$ and its canonical mesh $O_i$:

\begin{equation}
    (T_i, O_i) = F_\mathrm{AR}(P_i, M_i, \mathcal{F}_\mathrm{img}, P_\mathrm{scene})
\end{equation}

After processing all instances, the final scene reconstruction is obtained by transforming each canonical mesh into the scene coordinate frame $\mathcal{S} = \{ T_i O_i \}_{i=1}^N$.

We adopt EdgeRunner~\cite{tang2024edgerunner} and BPT~\cite{weng2025scaling} as our base models, both of which are autoregressive mesh generators designed for object-level, shape-conditioned generation. In their original formulations, a point-cloud encoder processes \textit{complete object point clouds} and produces conditioning tokens for the Transformer decoder to autoregressively generate mesh sequences. However, in single-view scene reconstruction, objects are only partially observed due to occlusions, and their global poses within the scene must also be inferred. In the following sections, we describe how we repurpose them for the single-view setting by (1) adapting the point-cloud encoder to incorporate appearance features from an image encoder, (2) injecting global scene context to compensate for missing geometry, and (3) predicting object poses within the same autoregressive framework.

\subsection{Repurposing the Point-Cloud Encoder}\label{sec:pc_encoder}

\paragraph{Injecting Pixel-Aligned Image Features}
The original point-cloud encoder used in EdgeRunner and BPT operates solely on point coordinates, without leveraging the rich appearance cues present in image features. To support single-view reconstruction, where objects are often partially observed, we augment the encoder with direct multi-modal fusion between geometry and pixel-aligned image features.

Given an instance point cloud $P_i$ and camera intrinsics $K$, each 3D point $p$ is projected onto the image plane to obtain its corresponding pixel $\mathrm{Proj}(K, p)=(u,v)$ on the global feature map $\mathcal{F}_{\mathrm{img}}$, establishing a point-pixel correspondence. For each such pair, the encoder $\mathcal{E}_{\mathrm{pc}}$ concatenates the geometric feature $\mathbf{f}_p^{\mathrm{pc}}$ with the aligned image feature $\mathbf{f}_p^{\mathrm{img}}=\mathcal{F}_{\mathrm{img}}(u, v)$ to form the key-value inputs to a Transformer-based fusion block. A set of learnable query embeddings then aggregates these fused features into a compact latent code:
\begin{equation}
    \mathbf{z}_i = \mathcal{E}_{\mathrm{pc}}\!\left(\mathbf{f}_p^{\mathrm{pc}},\, \mathbf{f}_p^{\mathrm{img}}\right) \; \forall p \in P_i.
\end{equation}
This pixel-aligned design enables the autoregressive mesh generator to incorporate per-point appearance cues, enhancing robustness to occlusion and improving the completeness and global consistency of the reconstructed geometry.

\paragraph{Scene Context Aggregation}
\label{sec:context_agg}
Instead of normalizing each instance independently in its own canonical space, which discards global spatial relations, we first normalize the entire global point cloud $P_\mathrm{scene}$ and all instance point clouds $\{P_i\}_{i=1}^N$ into a unified scene coordinate frame. This preserves consistent spatial reference among all objects. The normalized instance point clouds are then fed into the pixel-aligned point cloud encoder, ensuring that all encoded features share a coherent spatial frame for subsequent context aggregation. From this encoder, we obtain a scene-level latent $\mathbf{z}_\mathrm{scene}$ and per-instance latent codes $\mathbf{z}_i$.

To incorporate global scene context, \eg, cues from nearby objects of similar category or geometry, and to further improve reconstruction quality, each object latent $\mathbf{z}_i$ attends to the scene-level latent via a cross-attention layer:
\begin{equation}
    \mathbf{z}_i^{\mathrm{agg}}
    = \mathrm{CrossAttn}\!\left(
        q = \mathbf{z}_i,\,
        k = \mathbf{z}_\mathrm{scene},\,
        v = \mathbf{z}_\mathrm{scene}
    \right),
\end{equation}
The resulting aggregated feature $\mathbf{z}_i^{\mathrm{agg}}$ enriches the instance representation with holistic scene cues, enabling more accurate pose estimation and geometry prediction.

\subsection{Tokenization}
\label{sec:tokenization}

As an autoregressive framework, our model represents both object poses and meshes as discrete tokens. We uniformly quantize the unit cube $[-1, 1]^3$ into $N$ bins along each axis. For EdgeRunner, each vertex is represented by three integer tokens \texttt{<x>}, \texttt{<y>}, \texttt{<z>}, while BPT replaces these with a \texttt{<block\_id>} and \texttt{<offset\_id>} pair through block-wise decomposition of the $N^3$ quantized grid.

\paragraph{Object Pose Tokenization} Following standard conventions in 3D detection~\cite{lazarow2025cubify}, we represent each object pose using a gravity-aligned 7-DoF bounding box (center, scale, yaw). Rather than introducing a separate vocabulary for pose parameters (\eg, the yaw angle), we reuse the vertex tokenization scheme by encoding the $8$ corner points of the bounding box (normalized with respect to the global normalization in \cref{sec:context_agg}). This yields lightweight pose sequences (24 tokens for EdgeRunner and 16 tokens for BPT), negligible compared to mesh sequences. Importantly, this vertex-based formulation enables complete vocabulary sharing with mesh tokenization, avoiding new token types while maintaining expressiveness.

At inference time, the pose sequence is decoded into $8$ bounding-box corners in the normalized scene coordinate frame. The subsequent mesh sequence is decoded in the local canonical space, where each object is normalized to a unit cube. To bridge these two spaces, we recover a local-to-global transformation using the decoded global-space corners as targets. Let $\mathbf{X}_\mathrm{local} \in \mathbb{R}^{8 \times 3}$ denote the canonical box corners and $\mathbf{X}_\mathrm{global} \in \mathbb{R}^{8 \times 3}$ denote the decoded global-space corners. We estimate the best-fit affine transformation $\mathbf{T} \in \mathbb{R}^{3 \times 4}$ by solving the linear least-squares problem:
\begin{equation}
{\mathbf{T}}^\star 
= \underset{\mathbf{T}}{\arg\min}
\big\|\, \mathbf{X}_\mathrm{global} - [\mathbf{X}_\mathrm{local}\;\mathbf{1}]\, \mathbf{T}^\top \big\|_2^2.
\end{equation}
The resulting transformation $\mathbf{T}^\star$ is interpreted as a gravity-aligned transform, and is applied to all vertices of the decoded canonical mesh, yielding the final object geometry in the global scene frame.

\paragraph{Object Mesh Tokenization}

For mesh sequences, we adopt the native tokenization strategy of each base model.

BPT uses a \emph{Blocked and Patchified Tokenization} scheme that partitions the 3D coordinate grid into blocks and aggregates spatially adjacent faces into compact patches. This achieves strong compression (ratio $\approx 0.26$ at resolution $128$) with a large, structured vocabulary of $40{,}960$ tokens.

EdgeRunner employs a \emph{Compact Mesh Tokenization} derived from the EdgeBreaker algorithm~\cite{rossignac1999edgebreaker}, traversing triangles via a half-edge structure to maximize vertex reuse. It attains a moderate compression ratio ($\approx 0.46$ at resolution $512$) with a smaller vocabulary of $518$ tokens, while preserving high geometric fidelity.

These two tokenization paradigms are complementary: BPT prioritizes aggressive sequence compression with a high-capacity vocabulary, whereas EdgeRunner emphasizes resolution and geometric detail with a more compact vocabulary and moderate compression. In all cases, meshes are normalized to a unit cube and vertex coordinates are discretized according to the respective quantization resolution. Our framework supports both without modification, demonstrating robustness to widely different tokenization designs.

\paragraph{Final Token Sequence}
For each object, the final autoregressive sequence is constructed as:
\begin{gather*}
\resizebox{\linewidth}{!}{
\texttt{<bos>},\; \texttt{[pose\_seq]},\; \texttt{<sep>},\; \texttt{[mesh\_seq]},\; \texttt{<eos>}
}
\end{gather*}
where \texttt{[pose\_seq]} and \texttt{[mesh\_seq]} denote the tokenized pose and mesh sequences, respectively.

\subsection{Training}
\label{sec:training}

Our autoregressive decoder is trained using a single next-token prediction objective. Given a token
sequence $S=(s_1,\dots,s_T)$ and aggregated latent $\mathbf{z}_{\mathrm{agg}}$, the training loss is
\begin{equation}
    \mathcal{L}_{\mathrm{ce}}
    = -\sum_{t=1}^{T} 
      \log p_\theta\!\left(s_t \mid s_{<t},\, \mathbf{z}_{\mathrm{agg}}\right),
\end{equation}
where the model predicts each token conditioned on all preceding tokens and the fused point-cloud latent augmented with pixel-aligned image features and global scene context.

As illustrated in \cref{fig:pipeline}, the model autoregressively generates both the pose tokens and the mesh tokens within a single unified sequence. This joint formulation allows the decoder to learn instance geometry and global layout estimation simultaneously, enabling pose reasoning to benefit from geometry cues and vice versa.

%% file: sec/4_experiments.tex
\section{Experiments}

\subsection{Settings}

\input{tables/quantitative}

\paragraph{Datasets}
\label{sec:datasets}

We conduct experiments on synthetic and real-world datasets.
For training, we use the synthetic indoor dataset 3D-FRONT~\cite{3DFRONT}, adopting the preprocessed version provided by InstPIFu~\cite{InstPIFu}. Since the raw 3D-FRONT meshes are high-poly, we apply planar decimation to all object assets to obtain lightweight, artist-compatible meshes suitable for autoregressive generation. Additional preprocessing details are provided in the supplementary material. 3D-FRONT contains over 16K object meshes sourced from 3D-FUTURE~\cite{3DFuture}, along with scene layouts, RGB images, depth maps, and instance segmentation masks.

Following the standard protocol, our training split consists of 22{,}673 scene images. For evaluation on synthetic data, we use the test subset curated by DepR~\cite{Zhao_2025_ICCV_DepR}, which includes 100 scenes for object-level and 156 scenes for scene-level evaluation.

To demonstrate real-world generalization, we additionally evaluate our model on images from Pix3D~\cite{sun2018pix3d}, Matterport3D~\cite{chang2017matterport3d}, and ScanNet~\cite{dai2017scannet}. For these inputs, we employ Perspective Fields~\cite{jin2023perspective} to estimate camera intrinsics and pitch angles.

\paragraph{Implementation Details}
\label{sec:implementation}

For 2D visual priors, we follow DepR~\cite{Zhao_2025_ICCV_DepR} and employ off-the-shelf models: Grounded-SAM~\cite{ren2024grounded} for instance segmentation, Depth Pro~\cite{bochkovskii2024depth} for monocular depth estimation, and DINOv2 with register tokens~\cite{oquab2023dinov2,darcetvision} as our image feature encoder.

For back-projected point clouds, we adopt the native sampling densities of each base mesh generative model: BPT-based models use 4{,}096 points per object, whereas EdgeRunner-based models use 8{,}192 points. For the global scene representation, we uniformly sample 16{,}384 points.

All point clouds (partial object-level and full scene-level) and object poses are normalized to a unit cube. We apply random augmentation during training, including a rotation along the vertical axis in the range $[-45^\circ, 45^\circ]$, scaling in $[0.75, 1]$, and shift in $[0, 0.2]$. We additionally jitter depth values by up to 0.02 to account for inaccuracies in monocular depth estimation. Object meshes are normalized to a unit cube in their respective canonical space.

We train all models on 8 NVIDIA H100 GPUs using AdamW with a learning rate of $1\times10^{-4}$, 500 warm-up iterations, and cosine decay. The BPT-based variant converges in roughly 18 hours, while the EdgeRunner-based variant requires around 2 days due to its substantially longer token sequence length.

\paragraph{Evaluation Metrics}
\label{sec:metrics}
We evaluate our method using Chamfer Distance (CD) and F-Score, following standard practice in single-view reconstruction~\cite{InstPIFu, zhou2024zero,Zhao_2025_ICCV_DepR}. Unless otherwise noted, we use an F-Score threshold of 0.002. Each reconstructed mesh is uniformly sampled into 10k points prior to metric computation.

At the object level, we normalize predicted and ground-truth meshes to a unit cube and compute CD and F-Score to measure the geometric fidelity of individual objects.

At the scene level, we first assemble all predicted instances using their estimated poses. The composed scene, formed by placing each generated mesh into its predicted bounding box, remains in the normalized scene space described in \cref{sec:context_agg}. For fair comparison, we apply a global scale and translation to align the predicted scene with the ground-truth scene, which preserves its original metric scale and coordinate frame. Following DeepPriorAssembly~\cite{zhou2024zero}, we additionally report the single-direction Chamfer Distance (CD-S), which emphasizes reconstruction completeness while ignoring empty background regions.

\begin{figure*}
    \centering
    \includegraphics[width=\linewidth]{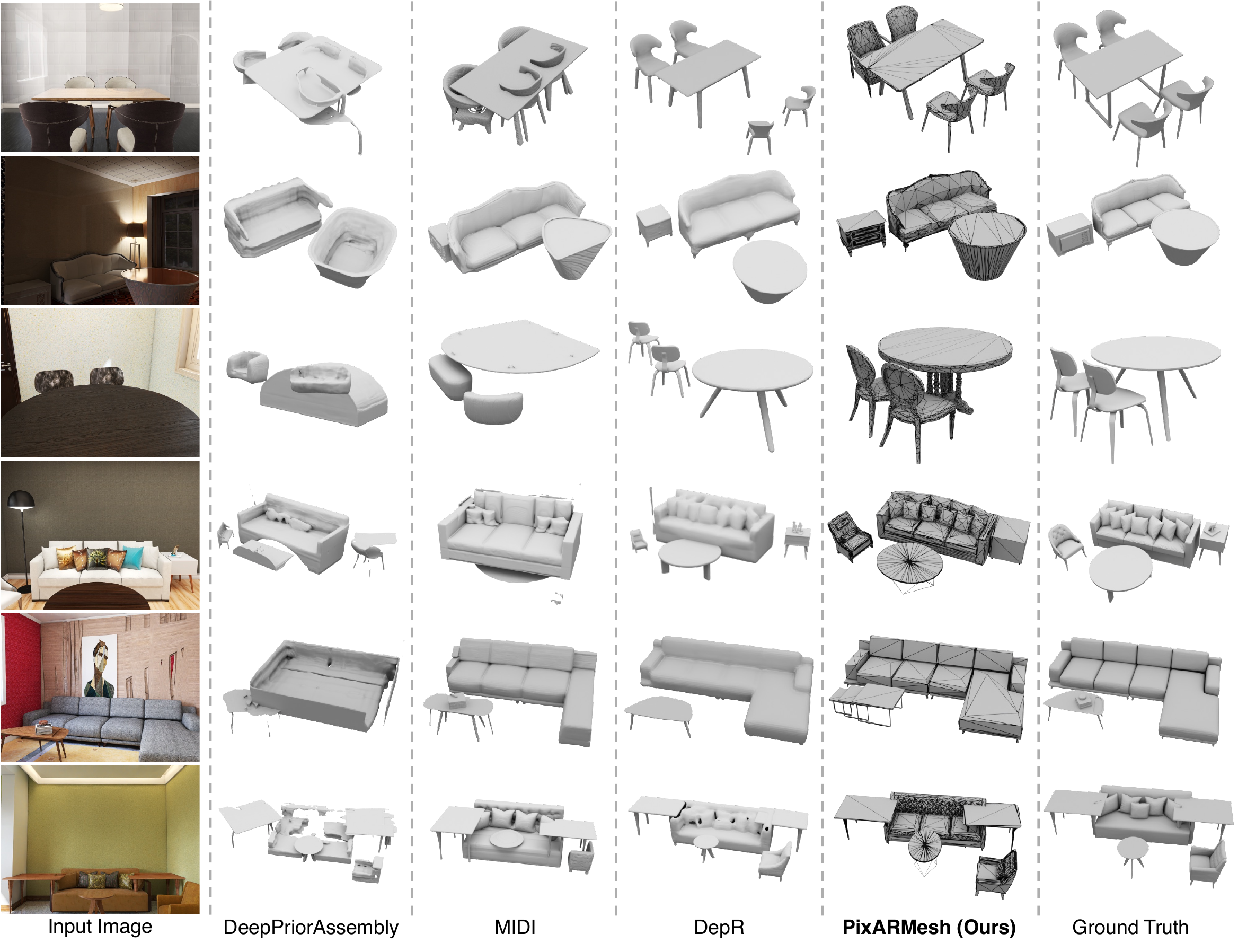}
    \caption{Qualitative comparisons on the 3D-FRONT~\cite{3DFRONT} dataset. For \OURS, we also show the mesh wireframe to highlight geometric quality.}
    \label{fig:qualitative_synth}
\end{figure*}

\begin{figure*}
    \centering
    \includegraphics[width=\linewidth]{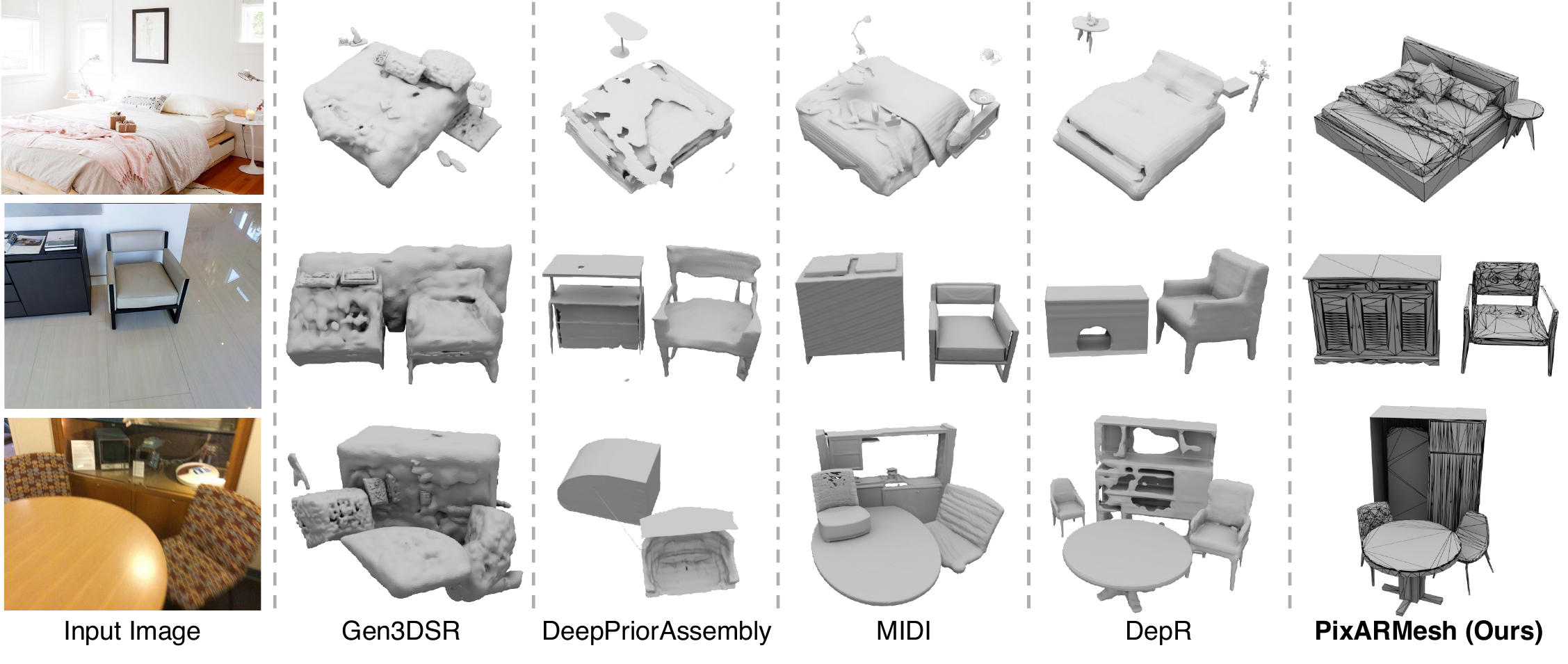}
    \caption{Qualitative results on real images from Pix3D~\cite{sun2018pix3d}, Matterport3D~\cite{chang2017matterport3d}, and ScanNet~\cite{dai2017scannet} datasets.}
    \label{fig:qualitative_real}
\end{figure*}

\subsection{Main Results}
\paragraph{Quantitative Results}
\cref{tab:quantitative} reports quantitative comparisons on the synthetic 3D-FRONT~\cite{3DFRONT} dataset.
We benchmark \OURS against representative single-view scene reconstruction approaches, including diffusion-based methods such as DepR~\cite{Zhao_2025_ICCV_DepR} and MIDI~\cite{huang2025midi}, feed-forward reconstruction frameworks such as InstPIFu~\cite{InstPIFu}, and holistic scene methods such as Uni-3D~\cite{zhang2023uni}.
Because holistic models do not explicitly generate individual object meshes, object-level metrics are not applicable.

Our method achieves highly competitive performance at both the object and scene levels. At the object level, \OURS achieves the second-best performance among all approaches, with F-Score comparable to diffusion-based SDF models. Unlike SDF-based pipelines that require Marching Cubes to extract dense iso-surfaces, our approach directly produces compact, artist-ready meshes with only a few thousand faces per instance while maintaining comparable geometric precision. Further statistics on face counts are provided in the supplementary material.
At the scene level, our method achieves state-of-the-art performance across all reported metrics. We attribute this to our unified autoregressive framework that jointly predicts object geometry and pose, leveraging our pixel-aligned point cloud encoder and scene-level context aggregation for coherent full-scene reconstruction.
We also observe that the EdgeRunner-based variant delivers stronger reconstruction performance than the BPT-based variant.

\paragraph{Qualitative Results}
We present qualitative comparisons on the synthetic 3D-FRONT~\cite{3DFRONT} dataset in \cref{fig:qualitative_synth} and on real-world images in \cref{fig:qualitative_real}.

Across both synthetic and real settings, \OURS produces geometrically coherent scene reconstructions, capturing object shapes and spatial arrangements that generally correspond to the input images. Owing to the native artist-ready mesh representation, \OURS yields meshes with clear edges and well-defined structural boundaries while maintaining smooth surface continuity, leading to cleaner shapes compared to prior approaches.

On real-world images, \OURS shows reasonable generalization and can reconstruct indoor environments with practical and interpretable geometry, despite being trained primarily on synthetic data.

\subsection{Ablation Studies}
We conduct ablations on the 3D-FRONT~\cite{3DFRONT} dataset to evaluate the effectiveness of \OURS{}.
Our analysis focuses on two aspects: (1) pipeline design -- examining the contribution of joint pose-mesh modeling and each proposed component, and (2) error analysis -- investigating the impact of upstream perception errors on overall scene reconstruction performance. We use EdgeRunner-based variant with pixel-aligned image features and scene context aggregation, unless otherwise stated.

\input{tables/ab_joint_model}

\paragraph{Necessity of Joint Pose-Mesh Modeling}
We construct additional baselines to validate the effectiveness of the proposed joint modeling of object poses and meshes introduced in \cref{sec:tokenization}. The results are summarized in \cref{tab:ab_joint_model}.

The \emph{EdgeRunner-FT} baseline is a fine-tuned EdgeRunner that reconstructs objects without layout conditions and composes scenes using the optimization strategy from DepR~\cite{Zhao_2025_ICCV_DepR}. It notably underperforms \OURS at the scene level, highlighting that explicit layout prediction is fundamentally more effective than post-hoc optimization.

The \emph{Two-stage} baseline decouples layout prediction and layout-conditioned mesh generation into separate models. Despite doubling the parameter count, it yields inferior performance across both scene-level and object-level metrics. In particular, joint modeling improves object reconstruction quality (CD $4.75 \rightarrow 4.04$, F-Score $80.85 \rightarrow 82.27$), suggesting that geometry generation benefits from being jointly optimized with pose prediction. While the two-stage design treats layout as a static condition, breaking the reasoning chain between localization and geometry, our unified formulation allows mesh generation to emerge as a coherent continuation of spatial reasoning.

\input{tables/ab_design}

\paragraph{Point-cloud Encoder Design}
To validate our repurposed point-cloud encoder, we report results with predicted depth and ground-truth masks in \cref{tab:ab_design}; additional results for the BPT-based variant are provided in the supplementary.

Removing the pixel-aligned image features causes the largest performance drop in both object and scene-level metrics. While using image features alone yields the highest object-level fidelity and the lowest scene-level Chamfer Distance, it falls short of the full model in terms of scene-level F-Score. These results highlight the importance of pixel-aligned image features for geometric accuracy and the synergy provided by scene-level context for overall reconstruction completeness.

\input{tables/ab_error_obj}

\paragraph{Object-Level Error Analysis}
To assess the upper-bound potential of \OURS and the impact of upstream perception errors, we evaluate it using ground-truth depth and layout for object-level reconstruction. As shown in \cref{tab:ablation_obj_recon}, replacing estimated depth with ground-truth depth provides the most significant boost in performance, reducing the Chamfer Distance from 4.13 to 3.04 and increasing the F-Score from 81.64\% to 86.66\%. While providing ground-truth layout alone yields more modest gains (82.27\% F-Score), the combination of both ground-truth depth and layout achieves the highest geometric fidelity with an F-Score of 87.19\%. These results indicate that while \OURS is robust to imperfect inputs, accurate depth and layout estimation offer essential guidance for generating the highest-quality mesh sequences.

\input{tables/ab_error_scene}

\paragraph{Scene-Level Error Analysis}
To isolate the influence of imperfect 2D visual priors, we evaluate \OURS using various combinations of ground-truth (GT) inputs. As shown in \cref{tab:ablation_error_scene}, providing GT segmentation yields the most substantial single-module improvement, boosting the scene-level F-Score from 33.55\% to 46.15\%. This sensitivity suggests that corrupted instance masks, which may lead to missing objects or fragmented point clouds, represent the primary bottleneck for scene-level reconstruction.

The results further indicate that the model exhibits robustness to monocular depth inaccuracies. Incorporating GT depth alone improves the F-Score from 33.55\% to 39.04\%, but its impact is less pronounced compared to the gains from idealized segmentation. Ultimately, the system achieves its peak performance with an F-Score of 68.48\% when all inputs are provided by an oracle. This demonstrates that while \OURS handles imperfect perception well, it scales effectively as upstream modules improve.

%% file: tables/quantitative.tex
\begin{table*}[!ht]
\centering
\setlength{\tabcolsep}{1em}
\resizebox{\linewidth}{!}{
\begin{tabular}{lcccccc}
\toprule

\multirow{2}{*}[-2pt]{\textbf{Method}} & \multicolumn{3}{c}{\textbf{Scene-level}} & \multicolumn{2}{c}{\textbf{Object-level}} \\

\cmidrule(lr){2-4} \cmidrule(lr){5-6}
    & \textbf{CD ($\times 10^{-3}$, $\downarrow$)} & \textbf{CD-S ($\times 10^{-3}$, $\downarrow$)} & \textbf{F-Score ($\%$, $\uparrow$)} & \textbf{CD ($\times 10^{-3}$, $\downarrow$)} & \textbf{F-Score ($\%$, $\uparrow$)} \\

\midrule

\rowcolor[RGB]{245,245,245} \multicolumn{6}{c}{SDF-based} \\

InstPIFu~\cite{InstPIFu}   & 213.4 & 124.9 & 13.72 & 44.74 & 29.63 \\
Uni-3D~\cite{zhang2023uni} & 218.3 & 113.3 & 12.99 & --- & --- \\
Gen3DSR~\cite{dogaru2024generalizable}& 222.4 & 137.5 & 13.52 & 9.74 & 31.42 \\
DeepPriorAssembly~\cite{zhou2024zero}& 191.8 & 76.2 & 16.72 & 20.13 & 27.83 \\
MIDI~\cite{huang2025midi}  & 156.3 & 79.3 & 24.83 & 6.71 & 72.69 \\
DepR~\cite{Zhao_2025_ICCV_DepR}  & 153.2 & 56.4 & 25.00 & \textbf{2.57} & \textbf{89.66} \\
\midrule

\rowcolor[RGB]{245,245,245} \multicolumn{6}{c}{Mesh-based} \\
\textbf{\OURS-EdgeRunner (Ours)} & 98.8 & 49.1 & \textbf{33.55} & \underline{4.04} & \underline{82.27} \\
\textbf{\OURS-BPT (Ours)} & \textbf{98.4} & \textbf{47.6} & 32.26 & 4.57 & 80.30 \\
\bottomrule
\end{tabular}
}
\caption{Qualitative comparison with state-of-the-art methods on the 3D-FRONT~\cite{3DFRONT} dataset. Following DepR~\cite{Zhao_2025_ICCV_DepR} and DeepPriorAssembly~\cite{zhou2024zero}, we report object- and scene-level Chamfer Distance (CD; lower is better) and F-Score (higher is better). We additionally include the single-direction Chamfer Distance (CD-S) to account for missing instances.}
\vspace{-1.5em}
\label{tab:quantitative}
\end{table*}

%% file: tables/ab_joint_model.tex
\begin{table}[htbp]
    \centering
    \resizebox{\linewidth}{!}{
    \setlength{\tabcolsep}{.2em}
    \begin{tabular}{lcccccc}
    \toprule
    \multirow{3}{*}{\textbf{Method}}
    & \multicolumn{3}{c}{\textbf{Scene-level}} 
    & \multicolumn{2}{c}{\textbf{Object-level}} \\
    & \textbf{CD}
    & \textbf{CD-S}
    & \textbf{F-Score}
    & \textbf{CD} 
    & \textbf{F-Score} \\
    & \textbf{($\times 10^{-3}$, $\downarrow$)}
    & \textbf{($\times 10^{-3}$, $\downarrow$)}
    & \textbf{($\%$, $\uparrow$)} & \textbf{($\times 10^{-3}$, $\downarrow$)} & \textbf{($\%$, $\uparrow$)} \\
    \midrule
    Two-stage                           &  99.8        &  50.6          & 33.32          & 4.75         &  80.85 \\
    EdgeRunner-FT  &  119.8       &  48.0          & 27.81          & 4.75         &  80.57 \\
    \textbf{\OURS (Ours)}               & \textbf{98.8} & \textbf{49.1} & \textbf{33.55} & \textbf{4.04} & \textbf{82.27} \\
    \bottomrule
    \end{tabular}
    }
    \caption{Ablation study on joint pose-mesh modeling.}
    \label{tab:ab_joint_model}
\end{table}

%% file: tables/ab_design.tex
\begin{table}[htbp]
    \centering
    \resizebox{\linewidth}{!}{
    \setlength{\tabcolsep}{.3em}
    \begin{tabular}{cccccccc}
    \toprule
    \multirow{3}{*}{\textbf{Img Feat}}
    & \multirow{3}{*}{\textbf{Ctx Agg}}
    & \multicolumn{3}{c}{\textbf{Scene-level}} 
    & \multicolumn{2}{c}{\textbf{Object-level}} \\
    &
    & \textbf{CD}
    & \textbf{CD-S}
    & \textbf{F-Score}
    & \textbf{CD} 
    & \textbf{F-Score} \\
    &
    & \textbf{($\times 10^{-3}$, $\downarrow$)}
    & \textbf{($\times 10^{-3}$, $\downarrow$)}
    & \textbf{($\%$, $\uparrow$)} & \textbf{($\times 10^{-3}$, $\downarrow$)} & \textbf{($\%$, $\uparrow$)} \\
    \midrule
               &            & 57.78  &  23.75  &  41.02  & 5.29  & 77.22  \\
               & \cmark & 55.44  &  27.16  &  42.84  & 5.56  & 78.14  \\
    \cmark &            & \textbf{39.30}  &  \textbf{16.12}  &  44.67  & \textbf{3.64}  & \textbf{84.64}  \\
    \cmark & \cmark & 39.88  &  18.52  &  \textbf{46.15}  & 4.04  & 82.27  \\
    \bottomrule
    \end{tabular}
    }
    \caption{Ablation studies on our point-cloud encoder design. \emph{Img Feat}, \emph{Ctx Agg} denote pixel-aligned image features and scene context aggregation, respectively.}
    \label{tab:ab_design}
\end{table}

%% file: tables/ab_error_obj.tex
\begin{table}[htbp]
    \centering
    \setlength{\tabcolsep}{2em}
    \resizebox{\linewidth}{!}{
    \begin{tabular}{cccc}
    \toprule
    \multirow{2}{*}{\textbf{GT Depth}} & \multirow{2}{*}{\textbf{GT Layout}} & \textbf{CD} & \textbf{F-Score} \\
     & & \textbf{($\times 10^{-3}$, $\downarrow$)} & \textbf{($\%$, $\uparrow$)} \\
    \midrule
                &            & 4.13  & 81.64 \\
    \cmark  &            & 3.04  & 86.66 \\
                & \cmark & 4.04  & 82.27 \\
    \cmark  & \cmark & \textbf{2.93}  & \textbf{87.19} \\
    \bottomrule
    \end{tabular}
    }
    \caption{Effects of depth and layout in object-level metrics.}
    \label{tab:ablation_obj_recon}
\end{table}

%% file: tables/ab_error_scene.tex
\begin{table}[htbp]
    \centering
    \setlength{\tabcolsep}{1em}
    \resizebox{\linewidth}{!}{
    \begin{tabular}{cccccc}
    \toprule
    \multicolumn{3}{c}{\textbf{GT Inputs}} & \textbf{CD} & \textbf{CD-S} & \textbf{F-Score} \\
    \textbf{Depth} & \textbf{Segm} & \textbf{Layout} & \textbf{($\times 10^{-3}$, $\downarrow$)} & \textbf{($\times 10^{-3}$, $\downarrow$)} & \textbf{($\%$, $\uparrow$)} \\
    \midrule
                   &            &          &  98.84  &  49.08  & 33.55  \\
        \cmark &            &          &  95.69  &  48.18  & 39.04  \\
                   & \cmark &          &  39.88  &  18.52  & 46.15  \\
                   & \cellcolor{gray!15}\cmark & \cellcolor{gray!15}\cmark & 16.30  &  4.40  &  63.91  \\
        \cmark & \cmark &          &  25.07  &  10.05  & 59.63  \\
        \cmark & \cellcolor{gray!15}\cmark & \cellcolor{gray!15}\cmark & \textbf{13.25}  & \textbf{3.52}  &  \textbf{68.48}  \\
    \bottomrule
    \end{tabular}
    }
    \caption{Effects of upstream (depth, segmentation, and layout) errors in scene-level metrics. Note that ground-truth layout implies ground-truth segmentation.}
    \label{tab:ablation_error_scene}
    \vspace{-0.5em}
\end{table}

%% file: sec/5_conclusion.tex
\section{Conclusion}

We presented \OURS, an autoregressive framework for single-view indoor scene reconstruction. By repurposing object-level mesh generative models with pixel-aligned image features and scene-level context aggregation, \OURS jointly predicts object pose and geometry, producing coherent full-scene reconstructions without relying on SDFs or post-hoc layout optimization. Our method achieves competitive object-level accuracy and state-of-the-art scene-level performance, notably while generating compact, artist-ready meshes. Extensive experiments and ablation studies highlight the effectiveness of our design and its applicability to real-world inputs, demonstrating the promise of autoregressive mesh generation as a viable alternative to conventional SDF-based pipelines.

%% file: sec/acknowledgment.tex
\section*{Acknowledgment} This work is supported by NSF award IIS-2127544 and NSF award IIS-2433768. We thank Lambda, Inc. for their compute resource help.

%% file: sec/supplementary.tex
\section{Additional Implementation Details}

\subsection{Data Pre-processing}

We pre-process meshes from the 3D-FRONT~\cite{3DFRONT} dataset to ensure they are suitable for autoregressive tokenization.

\begin{itemize}
\item \textbf{Vertex Merging} We merge nearby vertices using a minimum spatial resolution determined by the quantization level $q \in {128, 256, 512, 1024}$, where the minimum resolution is $1/q$.
\item \textbf{Mesh Simplification} We apply planar decimation followed by quadric-error-based edge-collapse decimation with target face counts of 800, 2000, and 4000.
\item \textbf{Quality Selection} We compute the Hausdorff distance between the simplified meshes and the originals, thresholding with an empirical value $\tau = 0.01$. If the distance is below $\tau$, we choose the mesh with fewer faces; otherwise, we select the quantization level and decimation setting that produce the lowest Hausdorff distance. This procedure balances compactness and geometric fidelity.
\end{itemize}

The resulting processed dataset contains meshes with an average of 1,809 faces per shape.

\subsection{Training Details}
To enhance layout prediction accuracy, we employ a two-stage training strategy:

\begin{itemize}
    \item \textbf{Layout Bootstrapping} We first train the model exclusively on the layout prediction prefix tokens. This stage consists of 100k iterations for the EdgeRunner-based variant and 30k iterations for the BPT-based model.
    \item \textbf{Joint Pose-Mesh Training} Building on the bootstrapped weights, we train the model on the full pose and mesh sequence. This final stage requires 30k iterations for EdgeRunner and 25k iterations for BPT.
\end{itemize}

Both stages utilize the AdamW optimizer with an initial learning rate of $1\times10^{-4}$, featuring a 500-step linear warmup followed by a cosine decay to $1\times10^{-5}$. Training is conducted with an effective batch size of 64. We set the optimizer hyperparameters to $\beta_1=0.9$ and $\beta_2=0.95$, and apply gradient clipping at a threshold of 1.0.

\section{Additional Results}

\subsection{More Qualitative Results on Real Images}

\begin{figure*}[!ht]
    \centering
    \includegraphics[width=\linewidth]{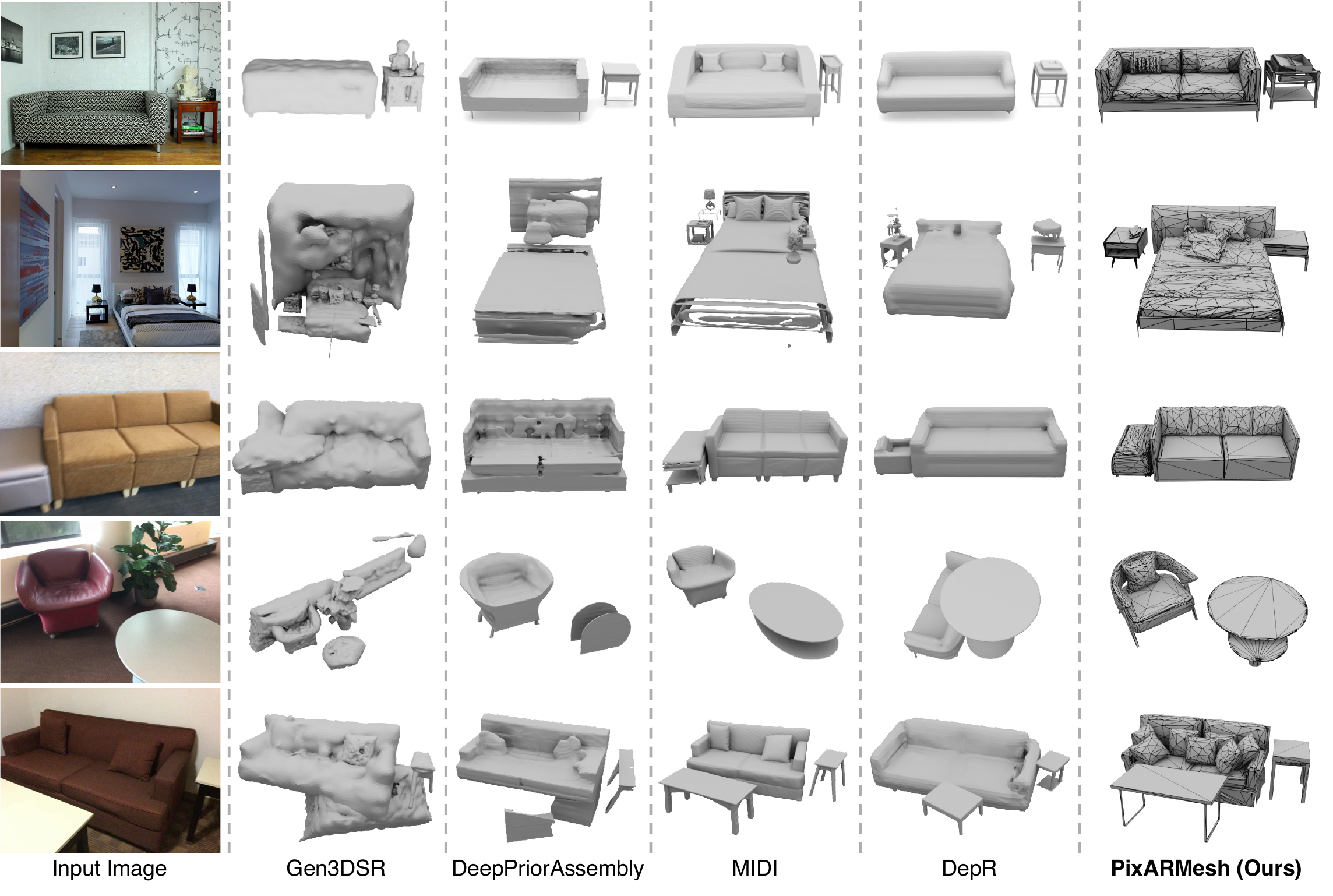}
    \caption{Additional qualitative results on real images from Pix3D~\cite{sun2018pix3d}, Matterport3D~\cite{chang2017matterport3d} and ScanNet~\cite{dai2017scannet}.}
    \label{fig:qualitative_real_supp}
\end{figure*}

We present additional qualitative results on real indoor scenes from Pix3D~\cite{sun2018pix3d}, Matterport3D~\cite{chang2017matterport3d} and ScanNet~\cite{dai2017scannet} in \cref{fig:qualitative_real_supp}. Across diverse environments and lighting conditions, \OURS generally achieves stronger layout alignment and produces coherent scene reconstructions, while retaining the advantages of compact, artist-ready mesh outputs.

\subsection{Face and Vertex Counts}

\input{tables/face_counts}

As shown in \cref{tab:face_count}, we report the average number of faces and vertices per reconstructed scene across different methods. InstPIFu~\cite{InstPIFu} and MIDI~\cite{huang2025midi} produce extremely dense outputs, requiring roughly 2M faces per scene. BUOL~\cite{chu2023buol}, Gen3DSR~\cite{dogaru2024generalizable}, and DeepPriorAssembly~\cite{zhou2024zero} reduce this to about 0.3M faces, though still far from mesh-efficient. In contrast, \OURS produces dramatically more compact meshes, \ie around 7--8k faces depending on the base model, while maintaining competitive geometric quality. This highlights the practical advantage of generating artist-ready, natively structured meshes rather than relying on iso-surface extraction.

\subsection{Object Pose (Layout) Accuracy}

\input{tables/layout}

Following MIDI~\cite{huang2025midi}, we evaluate scene layout accuracy using the 3D bounding-box IoU between predictions and ground truth. For this setting, we use ground-truth instance masks to isolate layout estimation from segmentation errors. Results are reported in \cref{tab:layout}. \OURS achieves higher layout accuracy than prior methods that rely on predicted depth, and we observe that the EdgeRunner-based variant consistently produces more accurate layouts than the BPT-based model.

\subsection{Runtime Analysis}

We measure inference runtime on a single NVIDIA A100 GPU and report the average per-scene latency in \cref{tab:runtime}. Due to its autoregressive decoding process, \OURS is inherently slower than feed-forward approaches (InstPIFu~\cite{InstPIFu}, Uni-3D~\cite{zhang2023uni}, BUOL~\cite{chu2023buol}) as well as the latent-diffusion-based DepR~\cite{Zhao_2025_ICCV_DepR}. Nonetheless, \OURS remains generally faster than other compositional pipelines such as Gen3DSR~\cite{dogaru2024generalizable}, while uniquely providing artist-ready native mesh outputs without needing iso-surface extraction.

\input{tables/runtime}

\subsection{BPT-Based Ablation Studies}

To further assess the generality of our point-cloud encoder designs, we conduct additional ablations using the BPT-based mesh generative model~\cite{weng2025scaling}, as summarized in \cref{tab:ab_design_bpt}. The trends mirror those observed with EdgeRunner~\cite{tang2024edgerunner}: incorporating scene-level context aggregation without pixel-aligned features slightly reduces object-level fidelity but improves global alignment in the assembled scene. Introducing image features provides a substantial boost in scene-level accuracy, and adding our scene-level context aggregation on top yields further gains. These results confirm that our proposed encoder designs remain effective across different mesh tokenizations and generative backbones.

\input{tables/ab_design_bpt}

%% file: tables/face_counts.tex
\begin{table}[ht]
\centering
\resizebox{\linewidth}{!}{
\setlength{\tabcolsep}{1em}
\begin{tabular}{lcc}
\toprule
\textbf{Method} & $|F|$ & $|V|$ \\
\midrule
  InstPIFu~\cite{InstPIFu} & 1,942,029 & 970,951 \\
  Uni-3D~\cite{zhang2023uni} & 141,130 & 70,844 \\
  BUOL~\cite{chu2023buol} & 55,493 & 27,750 \\
  Gen3DSR~\cite{dogaru2024generalizable} & 364,285 & 216,797 \\
  DeepPriorAssembly~\cite{zhou2024zero} & 250,875 & 125,420 \\
  MIDI~\cite{huang2025midi} & 1,936,248 & 967,886 \\ 
  DepR~\cite{Zhao_2025_ICCV_DepR} & 319,646 & 159,905 \\
  \textbf{\OURS-EdgeRunner (Ours)} & 7,110 & 4,253 \\
  \textbf{\OURS-BPT (Ours)} & 7,506 & 4,050 \\
\bottomrule
\end{tabular}
}
\caption{Face count $|F|$ and vertex count $|V|$ comparisons.}\label{tab:face_count}
\end{table}

%% file: tables/layout.tex
\begin{table}[ht]
\centering
\resizebox{\linewidth}{!}{
\setlength{\tabcolsep}{1em}
\begin{tabular}{lcc}
\toprule
\textbf{Method} & \textbf{GT Depth} & \textbf{Box IoU ($\%$)} \\
\midrule
  MIDI~\cite{huang2025midi}        & $-$          & 31.07 \\ 
  DepR~\cite{Zhao_2025_ICCV_DepR}  &              & 33.98 \\
  DepR~\cite{Zhao_2025_ICCV_DepR}  & $\checkmark$ & 36.67 \\
  \textbf{\OURS-EdgeRunner (Ours)} &              & 56.76 \\
  \textbf{\OURS-EdgeRunner (Ours)} & $\checkmark$ & 70.37 \\
  \textbf{\OURS-BPT (Ours)}        &              & 54.54 \\
  \textbf{\OURS-BPT (Ours)}        & $\checkmark$ & 65.01 \\
\bottomrule
\end{tabular}
}
\caption{Layout accuracy comparisons.}\label{tab:layout}
\end{table}

%% file: tables/runtime.tex
\begin{table}[ht]
\centering
\resizebox{0.7\linewidth}{!}{
\begin{tabular}{lc}
\toprule
\textbf{Method} & \textbf{Runtime} \\
\midrule
  InstPIFu~\cite{InstPIFu} & \qty{19.8}{\second} \\
  Uni-3D~\cite{zhang2023uni} & \qty{3.1}{\second} \\
  BUOL~\cite{chu2023buol} & \qty{6.5}{\second} \\
  Gen3DSR~\cite{dogaru2024generalizable} & \qty{15.5}{\minute} \\
  DeepPriorAssembly~\cite{zhou2024zero} & \qty{6.7}{\minute} \\
  MIDI~\cite{huang2025midi} & \qty{1.6}{\minute}  \\ 
  DepR~\cite{Zhao_2025_ICCV_DepR} & \qty{1.2}{\minute} \\
  \textbf{\OURS-EdgeRunner (Ours)} & \qty{4.5}{\minute} \\
  \textbf{\OURS-BPT (Ours)} & \qty{6.7}{\minute} \\
\bottomrule
\end{tabular}
}
\caption{Inference runtime comparisons.}\label{tab:runtime}
\end{table}

%% file: tables/ab_design_bpt.tex
\begin{table}[htbp]
    \centering
    \resizebox{\linewidth}{!}{
    \setlength{\tabcolsep}{.2em}
    \begin{tabular}{cccccccc}
    \toprule
    \multirow{3}{*}{\textbf{Img Feat}}
    & \multirow{3}{*}{\textbf{Ctx Agg}}
    & \multicolumn{3}{c}{\textbf{Scene-level}} 
    & \multicolumn{2}{c}{\textbf{Object-level}} \\
    &
    & \textbf{CD}
    & \textbf{CD-S}
    & \textbf{F-Score}
    & \textbf{CD} 
    & \textbf{F-Score} \\
    &
    & \textbf{($\times 10^{-3}$, $\downarrow$)}
    & \textbf{($\times 10^{-3}$, $\downarrow$)}
    & \textbf{($\%$, $\uparrow$)} & \textbf{($\times 10^{-3}$, $\downarrow$)} & \textbf{($\%$, $\uparrow$)} \\
    \midrule
               &            & 68.36  &  25.43  &  35.96  & 5.34  & 76.54  \\
               & \cmark & 56.80  &  23.49  &  37.98  & 5.85  & 76.31  \\
    \cmark &            & 57.03  &  23.70  &  41.51  & \textbf{4.25} & \textbf{81.25}  \\
    \cmark & \cmark & \textbf{49.82}  & \textbf{22.06} &  \textbf{42.18}  & 4.57  & 80.30  \\
    \bottomrule
    \end{tabular}
    }
    \caption{Ablation studies on our point-cloud encoder design with BPT-based model. \emph{Img Feat}, \emph{Ctx Agg} denote pixel-aligned image features and scene context aggregation, respectively.}
    \label{tab:ab_design_bpt}
\end{table}

%% file: main.bib
@String(ECCV= {Eur. Conf. Comput. Vis.})

@String(ECCV  = {ECCV})

@inproceedings{siddiqui2024meshgpt,
  title={Meshgpt: Generating triangle meshes with decoder-only transformers},
  author={Siddiqui, Yawar and Alliegro, Antonio and Artemov, Alexey and Tommasi, Tatiana and Sirigatti, Daniele and Rosov, Vladislav and Dai, Angela and Nie{\ss}ner, Matthias},
  booktitle={Proceedings of the IEEE/CVF Conference on Computer Vision and Pattern Recognition},
  pages={19615--19625},
  year={2024}
}

@inproceedings{
    chen2024meshanything,
    title={MeshAnything: Artist-Created Mesh Generation with Autoregressive Transformers},
    author={Yiwen Chen and Tong He and Di Huang and Weicai Ye and Sijin Chen and Jiaxiang Tang and Zhongang Cai and Lei Yang and Gang Yu and Guosheng Lin and Chi Zhang},
    booktitle={The Thirteenth International Conference on Learning Representations},
    year={2025},
    url={https://openreview.net/forum?id=KGZAs8VcOM}
}

@inproceedings{groueix2018papier,
  title={A papier-m{\^a}ch{\'e} approach to learning 3d surface generation},
  author={Groueix, Thibault and Fisher, Matthew and Kim, Vladimir G and Russell, Bryan C and Aubry, Mathieu},
  booktitle={Proceedings of the IEEE Conference on Computer Vision and Pattern Recognition},
  pages={216--224},
  year={2018}
}

@inproceedings{wang2018pixel2mesh,
  title={Pixel2mesh: Generating 3d mesh models from single rgb images},
  author={Wang, Nanyang and Zhang, Yinda and Li, Zhuwen and Fu, Yanwei and Liu, Wei and Jiang, Yu-Gang},
  booktitle={Proceedings of the European Conference on Computer Vision (ECCV)},
  pages={52--67},
  year={2018}
}

@inproceedings{dai2019scan2mesh,
  title={Scan2mesh: From unstructured range scans to 3d meshes},
  author={Dai, Angela and Nie{\ss}ner, Matthias},
  booktitle={Proceedings of the IEEE/CVF Conference on Computer Vision and Pattern Recognition},
  pages={5574--5583},
  year={2019}
}

@inproceedings{chen2020bsp,
  title={Bsp-net: Generating compact meshes via binary space partitioning},
  author={Chen, Zhiqin and Tagliasacchi, Andrea and Zhang, Hao},
  booktitle={Proceedings of the IEEE/CVF conference on computer vision and pattern recognition},
  pages={45--54},
  year={2020}
}

@inproceedings{nawrot2022hierarchical,
  title={Hierarchical transformers are more efficient language models},
  author={Nawrot, Piotr and Tworkowski, Szymon and Tyrolski, Micha{\l} and Kaiser, {\L}ukasz and Wu, Yuhuai and Szegedy, Christian and Michalewski, Henryk},
  booktitle={Findings of the Association for Computational Linguistics: NAACL 2022},
  pages={1559--1571},
  year={2022}
}

@article{chen2024meshxl,
  title={Meshxl: Neural coordinate field for generative 3d foundation models},
  author={Chen, Sijin and Chen, Xin and Pang, Anqi and Zeng, Xianfang and Cheng, Wei and Fu, Yijun and Yin, Fukun and Wang, Billzb and Yu, Jingyi and Yu, Gang and others},
  journal={Advances in Neural Information Processing Systems},
  volume={37},
  pages={97141--97166},
  year={2024}
}

@inproceedings{chen2025meshanything,
  title={Meshanything v2: Artist-created mesh generation with adjacent mesh tokenization},
  author={Chen, Yiwen and Wang, Yikai and Luo, Yihao and Wang, Zhengyi and Chen, Zilong and Zhu, Jun and Zhang, Chi and Lin, Guosheng},
  booktitle={Proceedings of the IEEE/CVF International Conference on Computer Vision},
  pages={13922--13931},
  year={2025}
}

@inproceedings{
weng2025pivotmesh,
title={PivotMesh: Generic 3D Mesh Generation via Pivot Vertices Guidance},
author={Haohan Weng and Yikai Wang and Tong Zhang and C. L. Philip Chen and Jun Zhu},
booktitle={The Thirteenth International Conference on Learning Representations},
year={2025},
url={https://openreview.net/forum?id=WAC8LmlKYf}
}

@inproceedings{zhang2025vertexregen,
  title={Vertexregen: Mesh generation with continuous level of detail},
  author={Zhang, Xiang and Siddiqui, Yawar and Avetisyan, Armen and Xie, Chris and Engel, Jakob and Howard-Jenkins, Henry},
  booktitle={Proceedings of the IEEE/CVF International Conference on Computer Vision},
  pages={12570--12580},
  year={2025}
}

@inproceedings{
    lei2025armesh,
    title={{ARM}esh: Autoregressive Mesh Generation via Next-Level-of-Detail Prediction},
    author={Jiabao Lei and Kewei Shi and Zhihao Liang and Kui Jia},
    booktitle={The Thirty-ninth Annual Conference on Neural Information Processing Systems},
    year={2025},
    url={https://openreview.net/forum?id=xlQ4QUB9VC}
}

@article{hao2024meshtron,
  title={Meshtron: High-fidelity, artist-like 3d mesh generation at scale},
  author={Hao, Zekun and Romero, David W and Lin, Tsung-Yi and Liu, Ming-Yu},
  journal={arXiv preprint arXiv:2412.09548},
  year={2024}
}

@inproceedings{lionar2025treemeshgpt,
  title={Treemeshgpt: Artistic mesh generation with autoregressive tree sequencing},
  author={Lionar, Stefan and Liang, Jiabin and Lee, Gim Hee},
  booktitle={Proceedings of the IEEE/CVF Conference on Computer Vision and Pattern Recognition},
  pages={26608--26617},
  year={2025}
}

@inproceedings{zhao2025deepmesh,
  title={Deepmesh: Auto-regressive artist-mesh creation with reinforcement learning},
  author={Zhao, Ruowen and Ye, Junliang and Wang, Zhengyi and Liu, Guangce and Chen, Yiwen and Wang, Yikai and Zhu, Jun},
  booktitle={Proceedings of the IEEE/CVF International Conference on Computer Vision},
  pages={10612--10623},
  year={2025}
}

@inproceedings{liu2025mesh,
    title={Mesh-{RFT}: Enhancing Mesh Generation via Fine-grained Reinforcement Fine-Tuning},
    author={Jian Liu and Jing Xu and Song Guo and Jing Li and Guojingfeng and Jiaao Yu and Haohan Weng and Biwen Lei and Xianghui Yang and Zhuo Chen and Fangqi Zhu and Tao Han and Chunchao Guo},
    booktitle={The Thirty-ninth Annual Conference on Neural Information Processing Systems},
    year={2025},
    url={https://openreview.net/forum?id=te2RsWcyQp}
}

@inproceedings{weng2025scaling,
  title={Scaling mesh generation via compressive tokenization},
  author={Weng, Haohan and Zhao, Zibo and Lei, Biwen and Yang, Xianghui and Liu, Jian and Lai, Zeqiang and Chen, Zhuo and Liu, Yuhong and Jiang, Jie and Guo, Chunchao and others},
  booktitle={Proceedings of the IEEE/CVF Conference on Computer Vision and Pattern Recognition},
  pages={11093--11103},
  year={2025}
}

@inproceedings{tang2024edgerunner,
    title={EdgeRunner: Auto-regressive Auto-encoder for Artistic Mesh Generation},
    author={Jiaxiang Tang and Zhaoshuo Li and Zekun Hao and Xian Liu and Gang Zeng and Ming-Yu Liu and Qinsheng Zhang},
    booktitle={The Thirteenth International Conference on Learning Representations},
    year={2025},
    url={https://openreview.net/forum?id=81cta3WQVI}
}

@InProceedings{3DFRONT,
    author    = {Fu, Huan and Cai, Bowen and Gao, Lin and Zhang, Ling-Xiao and Wang, Jiaming and Li, Cao and Zeng, Qixun and Sun, Chengyue and Jia, Rongfei and Zhao, Binqiang and Zhang, Hao},
    title     = {3D-FRONT: 3D Furnished Rooms With layOuts and semaNTics},
    booktitle = {Proceedings of the IEEE/CVF International Conference on Computer Vision},
    month     = {October},
    year      = {2021},
    pages     = {10933-10942}
}

@inproceedings{InstPIFu,
  title={Towards high-fidelity single-view holistic reconstruction of indoor scenes},
  author={Liu, Haolin and Zheng, Yujian and Chen, Guanying and Cui, Shuguang and Han, Xiaoguang},
  booktitle={European Conference on Computer Vision},
  pages={429--446},
  year={2022},
  organization={Springer}
}

@article{3DFuture,
  title={3d-future: 3d furniture shape with texture},
  author={Fu, Huan and Jia, Rongfei and Gao, Lin and Gong, Mingming and Zhao, Binqiang and Maybank, Steve and Tao, Dacheng},
  journal={International Journal of Computer Vision},
  volume={129},
  number={12},
  pages={3313--3337},
  year={2021},
  publisher={Springer}
}

@inproceedings{xu2024bayesian,
    author    = {Xu, Haiyang and Lei, Yu and Chen, Zeyuan and Zhang, Xiang and Zhao, Yue and Wang, Yilin and Tu, Zhuowen},
    title     = {Bayesian Diffusion Models for 3D Shape Reconstruction},
    booktitle = {Proceedings of the IEEE/CVF Conference on Computer Vision and Pattern Recognition},
    month     = {June},
    year      = {2024},
    pages     = {10628-10638}
}

@inproceedings{zhang2023uni,
  title={Uni-3d: A universal model for panoptic 3d scene reconstruction},
  author={Zhang, Xiang and Chen, Zeyuan and Wei, Fangyin and Tu, Zhuowen},
  booktitle={Proceedings of the IEEE/CVF International Conference on Computer Vision},
  pages={9256--9266},
  year={2023}
}

@article{dahnert2021panoptic,
  title={Panoptic 3d scene reconstruction from a single rgb image},
  author={Dahnert, Manuel and Hou, Ji and Nie{\ss}ner, Matthias and Dai, Angela},
  journal={Advances in Neural Information Processing Systems},
  volume={34},
  pages={8282--8293},
  year={2021}
}

@InProceedings{Zhao_2025_ICCV_DepR,
  author    = {Zhao, Qingcheng and Zhang, Xiang and Xu, Haiyang and Chen, Zeyuan and Xie, Jianwen and Gao, Yuan and Tu, Zhuowen},
  title     = {DepR: Depth Guided Single-view Scene Reconstruction with Instance-level Diffusion},
  booktitle = {Proceedings of the IEEE/CVF International Conference on Computer Vision},
  month     = {October},
  year      = {2025},
  pages     = {5722-5733}
}

@article{liu2021towards,
  title={Towards panoptic 3d parsing for single image in the wild},
  author={Liu, Sainan and Nguyen, Vincent and Gao, Yuan and Tripathi, Subarna and Tu, Zhuowen},
  journal={arXiv preprint arXiv:2111.03039},
  year={2021}
}

@article{alliegro2023polydiff,
  title={Polydiff: Generating 3d polygonal meshes with diffusion models},
  author={Alliegro, Antonio and Siddiqui, Yawar and Tommasi, Tatiana and Nie{\ss}ner, Matthias},
  journal={arXiv preprint arXiv:2312.11417},
  year={2023}
}

@inproceedings{nash2020polygen,
  title={Polygen: An autoregressive generative model of 3d meshes},
  author={Nash, Charlie and Ganin, Yaroslav and Eslami, SM Ali and Battaglia, Peter},
  booktitle={International conference on machine learning},
  pages={7220--7229},
  year={2020},
  organization={PMLR}
}

@inproceedings{dai2017scannet,
  title={Scannet: Richly-annotated 3d reconstructions of indoor scenes},
  author={Dai, Angela and Chang, Angel X and Savva, Manolis and Halber, Maciej and Funkhouser, Thomas and Nie{\ss}ner, Matthias},
  booktitle={Proceedings of the IEEE Conference on Computer Vision and Pattern Recognition},
  pages={5828--5839},
  year={2017}
}

@inproceedings{chang2017matterport3d,
  title={Matterport3D: Learning from RGB-D Data in Indoor Environments},
  author={Chang, Angel and Dai, Angela and Funkhouser, Thomas and Halber, Maciej and Niebner, Matthias and Savva, Manolis and Song, Shuran and Zeng, Andy and Zhang, Yinda},
  booktitle={2017 International Conference on 3D Vision},
  pages={667--676},
  year={2017},
  organization={IEEE Computer Society}
}

@article{rossignac1999edgebreaker,
  title={Edgebreaker: Connectivity compression for triangle meshes},
  author={Rossignac, Jarek},
  journal={IEEE Transactions on Visualization and Computer Graphics},
  volume={5},
  number={1},
  pages={47--61},
  year={1999},
  publisher={IEEE}
}

@article{oquab2023dinov2,
  title={{DINO}v2: Learning Robust Visual Features without Supervision},
    author={Maxime Oquab and Timoth{\'e}e Darcet and Th{\'e}o Moutakanni and Huy V. Vo and Marc Szafraniec and Vasil Khalidov and Pierre Fernandez and Daniel HAZIZA and Francisco Massa and Alaaeldin El-Nouby and Mido Assran and Nicolas Ballas and Wojciech Galuba and Russell Howes and Po-Yao Huang and Shang-Wen Li and Ishan Misra and Michael Rabbat and Vasu Sharma and Gabriel Synnaeve and Hu Xu and Herve Jegou and Julien Mairal and Patrick Labatut and Armand Joulin and Piotr Bojanowski},
    journal={Transactions on Machine Learning Research},
    issn={2835-8856},
    year={2024},
    url={https://openreview.net/forum?id=a68SUt6zFt},
    note={Featured Certification}
}

@inproceedings{bochkovskii2024depth,
    title={Depth Pro: Sharp Monocular Metric Depth in Less Than a Second},
    author={Alexey Bochkovskiy and Ama{\"e}l Delaunoy and Hugo Germain and Marcel Santos and Yichao Zhou and Stephan Richter and Vladlen Koltun},
    booktitle={The Thirteenth International Conference on Learning Representations},
    year={2025},
    url={https://openreview.net/forum?id=aueXfY0Clv}
}

@article{ren2024grounded,
  title={Grounded sam: Assembling open-world models for diverse visual tasks},
  author={Ren, Tianhe and Liu, Shilong and Zeng, Ailing and Lin, Jing and Li, Kunchang and Cao, He and Chen, Jiayu and Huang, Xinyu and Chen, Yukang and Yan, Feng and others},
  journal={arXiv preprint arXiv:2401.14159},
  year={2024}
}

@inproceedings{huang2025midi,
  title={Midi: Multi-instance diffusion for single image to 3d scene generation},
  author={Huang, Zehuan and Guo, Yuan-Chen and An, Xingqiao and Yang, Yunhan and Li, Yangguang and Zou, Zi-Xin and Liang, Ding and Liu, Xihui and Cao, Yan-Pei and Sheng, Lu},
  booktitle={Proceedings of the IEEE/CVF Conference on Computer Vision and Pattern Recognition},
  pages={23646--23657},
  year={2025}
}

@inproceedings{dogaru2024generalizable,
    title={Gen3{DSR}: Generalizable 3D Scene Reconstruction via Divide and Conquer from a Single View},
    author={Andreea Dogaru and Mert {\"O}zer and Bernhard Egger},
    booktitle={International Conference on 3D Vision 2025},
    year={2025},
    url={https://openreview.net/forum?id=FFlGtmV1Co}
}

@inproceedings{chu2023buol,
  title={Buol: A bottom-up framework with occupancy-aware lifting for panoptic 3d scene reconstruction from a single image},
  author={Chu, Tao and Zhang, Pan and Liu, Qiong and Wang, Jiaqi},
  booktitle={Proceedings of the IEEE/CVF Conference on Computer Vision and Pattern Recognition},
  pages={4937--4946},
  year={2023}
}

@article{zhou2024zero,
  title={Zero-shot scene reconstruction from single images with deep prior assembly},
  author={Zhou, Junsheng and Liu, Yu-Shen and Han, Zhizhong},
  journal={Advances in Neural Information Processing Systems},
  volume={37},
  pages={39104--39127},
  year={2024}
}

@incollection{lorensen1998marching,
  title={Marching cubes: A high resolution 3D surface construction algorithm},
  author={Lorensen, William E and Cline, Harvey E},
  booktitle={Seminal graphics: pioneering efforts that shaped the field},
  pages={347--353},
  year={1998},
  publisher = {Association for Computing Machinery}
}

@article{jun2023shap,
  title={Shap-e: Generating conditional 3d implicit functions},
  author={Jun, Heewoo and Nichol, Alex},
  journal={arXiv preprint arXiv:2305.02463},
  year={2023}
}

@inproceedings{liu2023zero,
  title={Zero-1-to-3: Zero-shot one image to 3d object},
  author={Liu, Ruoshi and Wu, Rundi and Van Hoorick, Basile and Tokmakov, Pavel and Zakharov, Sergey and Vondrick, Carl},
  booktitle={Proceedings of the IEEE/CVF International Conference on Computer Vision},
  pages={9298--9309},
  year={2023}
}

@article{zhang2024clay,
  title={Clay: A controllable large-scale generative model for creating high-quality 3d assets},
  author={Zhang, Longwen and Wang, Ziyu and Zhang, Qixuan and Qiu, Qiwei and Pang, Anqi and Jiang, Haoran and Yang, Wei and Xu, Lan and Yu, Jingyi},
  journal={ACM Transactions on Graphics},
  volume={43},
  number={4},
  pages={1--20},
  year={2024},
  publisher={ACM New York, NY, USA}
}

@inproceedings{lazarow2025cubify,
  title={Cubify anything: Scaling indoor 3d object detection},
  author={Lazarow, Justin and Griffiths, David and Kohavi, Gefen and Crespo, Francisco and Dehghan, Afshin},
  booktitle={Proceedings of the IEEE/CVF Conference on Computer Vision and Pattern Recognition},
  pages={22225--22233},
  year={2025}
}

@inproceedings{darcetvision,
  title={Vision Transformers Need Registers},
  author={Darcet, Timoth{\'e}e and Oquab, Maxime and Mairal, Julien and Bojanowski, Piotr},
  booktitle={The Twelfth International Conference on Learning Representations},
  year={2024}
}

@inproceedings{hong2023lrm,
    title={{LRM}: Large Reconstruction Model for Single Image to 3D},
    author={Yicong Hong and Kai Zhang and Jiuxiang Gu and Sai Bi and Yang Zhou and Difan Liu and Feng Liu and Kalyan Sunkavalli and Trung Bui and Hao Tan},
    booktitle={The Twelfth International Conference on Learning Representations},
    year={2024},
    url={https://openreview.net/forum?id=sllU8vvsFF}
}

@inproceedings{jin2023perspective,
  title={Perspective fields for single image camera calibration},
  author={Jin, Linyi and Zhang, Jianming and Hold-Geoffroy, Yannick and Wang, Oliver and Blackburn-Matzen, Kevin and Sticha, Matthew and Fouhey, David F},
  booktitle={Proceedings of the IEEE/CVF Conference on Computer Vision and Pattern Recognition},
  pages={17307--17316},
  year={2023}
}

@article{liu2023one,
  title={One-2-3-45: Any single image to 3d mesh in 45 seconds without per-shape optimization},
  author={Liu, Minghua and Xu, Chao and Jin, Haian and Chen, Linghao and Varma T, Mukund and Xu, Zexiang and Su, Hao},
  journal={Advances in Neural Information Processing Systems},
  volume={36},
  pages={22226--22246},
  year={2023}
}

@inproceedings{kuo2020mask2cad,
  title={Mask2cad: 3d shape prediction by learning to segment and retrieve},
  author={Kuo, Weicheng and Angelova, Anelia and Lin, Tsung-Yi and Dai, Angela},
  booktitle={European Conference on Computer Vision},
  pages={260--277},
  year={2020},
  organization={Springer}
}

@inproceedings{
    huang2025litereality,
    title={LiteReality: Graphic-Ready 3D Scene Reconstruction from {RGB}-D Scans},
    author={Zhening Huang and Xiaoyang Wu and Fangcheng Zhong and Hengshuang Zhao and Matthias Nie{\ss}ner and Joan Lasenby},
    booktitle={The Thirty-ninth Annual Conference on Neural Information Processing Systems},
    year={2025},
    url={https://openreview.net/forum?id=4JLZsmWBJf}
}

@inproceedings{liu2024one,
  title={One-2-3-45++: Fast single image to 3d objects with consistent multi-view generation and 3d diffusion},
  author={Liu, Minghua and Shi, Ruoxi and Chen, Linghao and Zhang, Zhuoyang and Xu, Chao and Wei, Xinyue and Chen, Hansheng and Zeng, Chong and Gu, Jiayuan and Su, Hao},
  booktitle={Proceedings of the IEEE/CVF Conference on Computer Vision and Pattern Recognition},
  pages={10072--10083},
  year={2024}
}

@inproceedings{sun2018pix3d,
  title={Pix3d: Dataset and methods for single-image 3d shape modeling},
  author={Sun, Xingyuan and Wu, Jiajun and Zhang, Xiuming and Zhang, Zhoutong and Zhang, Chengkai and Xue, Tianfan and Tenenbaum, Joshua B and Freeman, William T},
  booktitle={Proceedings of the IEEE conference on computer vision and pattern recognition},
  pages={2974--2983},
  year={2018}
}
